\documentclass{article}

\usepackage{arxiv}

\usepackage[utf8]{inputenc} 
\usepackage[T1]{fontenc}    
\usepackage{hyperref}       
\usepackage{url}            
\usepackage{booktabs}       
\usepackage{amsfonts}       
\usepackage{nicefrac}       
\usepackage{microtype}      
\usepackage{lipsum}		
\usepackage{graphicx}
\usepackage{natbib}
\usepackage{doi}
\usepackage{authblk}
\usepackage{caption} 
\usepackage{amsmath}
\usepackage{bbm}
\usepackage{multirow}
\usepackage{comment}
\usepackage{amsthm}

\newcommand{\E}{\mathbb{E}}

\newtheorem{property}{Property}

\begin{document}
\title{Deep offline reinforcement learning for real-world treatment optimization applications}


\author[1, *]{Mila Nambiar}
\author[1, 2, *]{Supriyo Ghosh}
\author[1]{Yu En Chan}
\author[1]{Priscilla Ong}
\author[3]{Yong Mong Bee}
\author[1]{Pavitra Krishnaswamy}

\affil[1]{\footnotesize Institute for Infocomm Research (I\textsuperscript{2}R), A\textsuperscript{*}STAR, Singapore}
\affil[2]{\footnotesize Microsoft, Bengaluru, Karnataka, India}
\affil[3]{\footnotesize Department of Endocrinology, Singapore General Hospital, Singapore}
\affil[*]{Both authors contributed equally to this research.}



\renewcommand{\shorttitle}{Nambiar and Ghosh, et al.}

\hypersetup{
pdftitle={Deep offline reinforcement learning for real-world treatment optimization applications},
pdfauthor={Milashini Nambiar, Supriyo Ghosh},
pdfkeywords={Offline reinforcement learning; dynamic treatment optimization; sepsis; type 2 diabetes; sampling; safety constraints},
}

\maketitle

\begin{abstract}
There is increasing interest in data-driven approaches for recommending optimal treatment strategies in many chronic disease management and critical care applications. Reinforcement learning methods are well-suited to this sequential decision-making problem, but must be trained and evaluated exclusively on retrospective medical record datasets as direct online exploration is unsafe and infeasible. Despite this requirement, the vast majority of treatment optimization studies use off-policy RL methods (e.g., Double Deep Q Networks (DDQN) or its variants) that are known to perform poorly in purely offline settings. Recent advances in offline RL, such as Conservative Q-Learning (CQL), offer a suitable alternative. But there remain challenges in adapting these approaches to real-world applications where suboptimal examples dominate the retrospective dataset and strict safety constraints need to be satisfied. In this work, we introduce a practical and theoretically grounded transition sampling approach to address action imbalance during offline RL training. We perform extensive experiments on two real-world tasks for diabetes and sepsis treatment optimization to compare performance of the proposed approach against prominent off-policy and offline RL baselines (DDQN and CQL). Across a range of principled and clinically relevant metrics, we show that our proposed approach enables substantial improvements in expected health outcomes and in accordance with relevant practice and safety guidelines.
\end{abstract}

\keywords{Offline reinforcement learning, dynamic treatment optimization, sepsis, type 2 diabetes, sampling, safety constraints}

\section{Introduction}
\label{sec:introduction}

Deep reinforcement learning (RL) has recently experienced a surge in popularity thanks to demonstrated successes in game playing (e.g. Atari and Go \citep{mnih2013playing,silver2017mastering}), and with AI bots (e.g. ChatGPT\footnote{https://openai.com/blog/chatgpt/}). Given its ability to learn from large real-world experience datasets, there is also immense excitement about the potential of deep RL for clinical decision support applications.  In many such applications, the objective is to leverage historical medical records containing information on patient characteristics, disease state evolution, treatment decisions and clinical outcomes; and learn treatment policies that will optimize clinical outcomes of interest. Notably, deep RL has been used for treatment optimization for a range of clinical conditions including sepsis, hypertension, type 2 diabetes, and cancer \citep{sepsisraghu, sepsiswisAIM, t2dmsdr, t2dmkNN, tseng2017}. However, unlike traditional game-playing or consumer-oriented applications on which deep RL methods have been developed and widely tested, treatment optimization applications are not amenable to learning through active interaction. This is due to critical safety concerns, which forbid direct online exploration of treatment alternatives on patients.

Offline RL (also known as batch RL \citep{lange2012batch}), an approach to learn from large, previously collected datasets without any interaction with an environment, is then ideally suited for treatment optimization applications. Yet, the treatment optimization literature has almost exclusively focused on traditional value-based off-policy RL methods, particularly Double Deep Q Networks (DDQN) \citep{ddqn} and its variants \citep{t2dmsdr, t2dmkNN, sepsisraghu, sepsislu, sepsispeng, sepsisyu, zhu2021type1rnn}. However, direct use of off-policy RL algorithms in an offline setting is known to perform poorly in general, due to issues with bootstrapping from out-of-distribution (OOD) actions \citep{kumar2019stabilizing} and overfitting to unseen actions \citep{agarwal2020optimistic,fu2019diagnosing}.  In other words, off-policy methods could overestimate the Q-values of unseen state-action pairs, and mistakenly select unacceptable or even unsafe actions. Recently proposed offline RL methods such as Conservative Q-Learning (CQL) and Model-based Offline Policy Optimization (MOPO) \citep{cql,yu2020mopo} address this overestimation problem by regularizing the Q-values for unseen actions during training and by lower-bounding value function estimates.

However, translating these advances in offline RL directly to real-world treatment optimization applications remains challenging. One key challenge is that medical record datasets reflect real-world clinical practice and hence contain a mixture of both optimal and suboptimal actions. In many cases, suboptimal treatments may be prescribed due to patient preferences, communication difficulties, time and resource restrictions, limitations in clinician experience, and/or inherent uncertainty in determining the best treatment strategy (e.g., due to conflicting clinical trial evidence) \citep{suboptimalexs, shah2005clinicalinertia}. Often, these practice barriers give rise to behavior policy distributions that are heavily imbalanced, with the frequency of suboptimal actions even outweighing the frequency of optimal actions. In this context, offline RL methods that severely penalize out-of-distribution actions may result in overly conservative policies. 

To address this challenge, we leverage sampling methodologies to adapt offline RL methods for scenarios where suboptimal examples dominate the retrospective dataset. Specifically, our approach samples historical records of transitions (i.e., state, action, and reward tuples) corresponding to each action without altering the transition probabilities, in order to increase the proportion of less frequently seen actions in the training data. We performed extensive experiments to compare performance of a popular off-policy RL baseline (DDQN) and a SOTA offline RL method (Conservative Q Learning) with and without our sampling approach on two real-world tasks for type 2 diabetes and sepsis treatment optimization. We assessed expected health outcomes via principled off-policy evaluations and characterized consistency with relevant practice and safety guidelines for the different methods.  

Our main contributions are summarized as follows:
\begin{itemize}
\item We demonstrate that CQL, a SOTA offline RL method, can be applied in real-world treatment optimization applications to make recommendations that are more aligned with clinical practice than DDQN, a popular off-policy RL method, while also improving expected health outcomes over DDQN and the standard of care (SoC).
\item We argue theoretically and demonstrate empirically that when an intuitive heuristic to strictly enforce safety constraints during policy execution is applied to CQL's and DDQN's recommendations, the relative improvement of CQL over DDQN extends to constrained recommendations.
\item We propose a transition sampling approach to address action imbalance, and show that this increases the likelihood of CQL selecting less frequently seen actions, while continuing to penalize value estimates for out-of-distribution actions.
\item Extensive experimental results for two real-world healthcare applications demonstrate that CQL with sampling substantially improves expected health outcomes over the SoC and CQL baselines, while ensuring high alignment with clinical practice.
\end{itemize} 

Our results suggest that offline RL, as opposed to off-policy RL, should be leveraged as a means of devising safe and clinically effective policies for treatment optimization problems.  

\section{Related Work}
\label{sec:related}
Our work focuses on deep offline RL methods for treatment optimization. We categorize existing relevant research into three threads: (a) Treatment optimization using RL; (b) Offline RL methods and their applications; and (c) Practical challenges associated with RL-based treatment optimization. 

\par{\bf{Treatment optimization using deep RL: }} The overall literature on RL for treatment optimization is comprehensively surveyed in \cite{healthcarerl2021}. Here, we review recent studies on deep reinforcement learning for treatment optimization applications. This literature has largely focused on optimizing management of complex syndromes in intensive inpatient settings \citep{sepsisraghu, raghusepsisb, sepsiswisAIM}, optimizing medication dosing in anesthesia and critical care \citep{anesth2020, ventsed2020}, and optimizing treatment choices for chronic diseases in outpatient settings \citep{ tseng2017, t2dmsdr, t2dmkNN}. Commonly, these works apply value-based off-policy deep RL algorithms such as DQN, DDQN, and variants (e.g., with dueling architecture, recurrent networks) on retrospective or batch datasets -- an approach which is known to suffer from distribution shift between the learned and behavior policies and overfit to unseen actions \citep{sepsisraghu,raghusepsisb,sepsispeng,sepsisyu,sepsislu,t2dmkNN, t2dmsdr}. 

\par{\bf{Offline RL:}}  The need to learn optimal policies in practical data-driven decision making scenarios has led to the development of offline RL algorithms. These algorithms are set up to learn effectively from retrospective data collected under some behavior policy, without any direct exploration or environmental interaction during training. We review recent works on deep offline RL. First, implicit constraint Q-Learning \citep{offlineRLimplicitconstraintyang} leverages imitation learning \citep{imitationlearningwang, imitationlearningchen} to address the overfitting problem by avoiding querying OOD samples. Second, value-based and policy-based offline RL algorithms (e.g., CQL \citep{cql}, MOPO \citep{yu2020mopo}, UWAC \citep{offlinerlwu}, Fisher-BRC \citep{Kostrikov2021OfflineRL}, COMBO \citep{yu2021combo}) prevent over-optimism by penalizing the learned value function for OOD actions with regularization during training. A third set of offline RL algorithms (e.g., BCQ \citep{bcq}, BEAR \citep{offlinebear}, BRAC \citep{behaviorregularizedofflinerl}) uses regularization to penalize deviations between the learned and behaviour policies. A few recent studies have explored offline RL algorithms for treatment optimization \citep{fatemisemimarkov,fatemideadendstates,representationlearningbcq}.  Among these, Fatemi \emph{et al.} \citep{fatemisemimarkov} proposed a modification to BCQ \citep{bcq} for a continuous-time semi-MDP setting. Further, Fatemi \emph{et al.} \citep{fatemideadendstates} proposed a method to identify states from which negative outcomes are unavoidable.  Finally, Killian \emph{et al.} \citep{representationlearningbcq} used state representation learning with discretized BCQ. However, translation of advances in offline RL to real-world treatment optimization applications is still nascent, and several practical challenges remain. 

\par{\bf{Practical challenges in RL-based treatment optimization:}} A key challenge in real-world clinical applications stems from action imbalance due to dominance of suboptimal, and often conservative, actions in the data.  The offline RL community is starting to recognize that overly conservative policies impede generalization and performance, and attempts to reduce conservativeness of CQL and variants are emerging. For example, the very recently proposed MCQ \citep{mildCQL} adapts CQL to actively train OOD actions. However, such strategies are not designed to directly address action imbalance in the retrospective data and cannot be generalized across offline RL methods. Inspired by sampling-based approaches to handle class imbalance in supervised learning \citep{Undersamplingkubat, overundersamplingling, underoversamplingsolberg, smote, adasyn}, we propose a transition sampling scheme to address action imbalance and demonstrate its ability to improve quality and relevance of the resulting treatment policies. 

\section{Background}
\label{sec:background}
In this section, we represent the treatment optimization problem within an RL framework and introduce baseline approaches such as Q-learning and DDQN. 
\subsection{Problem Formulation}
We consider a setting where the patient state evolves according to an underlying Markov Decision Process (MDP).  This MDP is defined by the tuple $(\mathcal{S}, \mathcal{A}, \mathcal{P}, r, \gamma)$, where $\mathcal{S}$ denotes the set of all possible states, $\mathcal{A}$ denotes the set of discrete permissible actions, $\mathcal{P}: \mathcal{S} \times \mathcal{A} \rightarrow \mathbf{P}(\mathcal{S})$ represents the transition function providing the next-state distribution after executing action $a\in \mathcal{A}$ in state $s\in \mathcal{S}$, $r: \mathcal{S} \times \mathcal{A} \rightarrow \mathbb{R}$ denotes a reward function providing the expected immediate reward for executing action $a$ in state $s$, and $\gamma \in [0,1]$ is a discount factor.  Let $T_i$ denote the treatment horizon length for patient $i$. At time $t$, for patient $i$, a clinician observes patient state $s_{i,t} \in \mathcal{S}$ and recommends a treatment, or action $a_{i,t}$ from a finite and discrete action set $\mathcal{A} = \{1, 2,..., A\}$.
The reward $r(s,a)$ is increasing in positive health outcomes (e.g., lab results within target range) and decreasing in negative health outcomes (e.g., mortality). The goal is to identify a treatment policy $\pi: \mathcal{S}\rightarrow \mathbf{P}(\mathcal{A})$ that chooses the action at each time step $t$ that maximizes expected cumulative reward as follows:
\begin{equation}
\begin{aligned}
\label{eq:rlopt}
\max_{\pi} \quad & \E \big[ \sum_{t'=t}^{T_i} \gamma^{t'-t}[r(s_{t',i}, \pi(s_{t',i}))]\big ]
\end{aligned}
\end{equation}
For some medical conditions, e.g., type 2 diabetes, feasible treatments are constrained by clinical practice and safety guidelines. This setting can be modeled as a constrained optimization problem:
\begin{equation}
\begin{aligned}
\label{eq:rloptconstr}
\max_{\pi} \quad & \E \big[ \sum_{t'=t}^{T_i} \gamma^{t'-t}[r(s_{t',i}, \pi(s_{t',i}))]\big ]\\
\textrm{s.t.} \quad & \E[C_j(s_{t, i}, \pi(s_{t',i}))] \leq c_j \hspace{0.2in} \forall j \in \{1,...,C\}  \\
\end{aligned}
\end{equation}
$C_j(s_{t, i}, \pi(s_{t',i}))$ denotes the type $j$ constraint violation cost at time $t$ for taking action $\pi(s_{t',i})$ at state $s_{t, i}$. $c_j$ is the threshold on cumulative constraint violation cost over the entire horizon for type $j$ constraints.  It can be set as 0 for treatment recommendation applications, to represent strict safety guidelines. 

We are specifically interested in the offline reinforcement learning setting, where the policy that solves Equation~\ref{eq:rlopt} must be learned solely from some retrospective dataset $\mathcal{D} = \{(s_{i,t}, a_{i,t}, r_{i,t}, s_{i,t+1})$, $\text{ for } i = 1,..., I; t = 1,...,T_i\}$ generated by a behavior policy $\pi_{\beta}(s)$.  This behavior policy corresponds to the SoC, and need not be optimal or strictly satisfy constraints due to real-world clinical practice challenges. As we are interested in comparing the performance of offline RL methods with off-policy RL methods, we begin by introducing the off-policy RL method, DDQN, which leverages the canonical RL algorithm, Q-learning. 

\subsection{Q-learning}
Our goal is to learn a (potentially randomized) policy $\pi: \mathcal{S}\rightarrow \mathbf{P}(\mathcal{A})$ that specifies the action to take in each state $s \in \mathcal{S}$. The value at state $s$ with respect to policy $\pi$ is estimated as the expected cumulative reward for executing $\pi$:
\begin{equation}
\label{eq:valueofpolicy}
V^\pi(s) = \E_\pi\big[\sum_{t=0}^T \gamma^t r(s_t,\pi(s_t))| s_0 = s\big]
\end{equation}

The goal is to learn an optimal policy $\pi^*$ that maximizes $V^{\pi}(s)$ for all $s$. In a finite MDP setting, tabular Q-learning can asymptotically learn $\pi^*$ by learning the optimal $Q$ function \citep{qthesis} $Q^{\pi^*}$, where $Q^{\pi^*}$ is obtained using the following Bellman equation:
\begin{equation}
Q^{\pi^*}(s,a) = r(s,a) + \gamma \E_{s'\sim \mathbf{P}(s'|s,a)}\big[\max_{a'} Q^{\pi^*}(s',a')\big]
\end{equation}
If $Q^{\pi^*}$ is known, the optimal action $a^*$ for a given state $s$ is computed by $a^* = \pi^*(s) = \arg\max_{a} Q^{\pi^*}(s,a)$.
$Q^{\pi^*}$ can be estimated by iteratively applying the following Bellman operator: 
\begin{equation}
\label{eq:bellmanop}
\mathcal{B}^kQ^k(s,a) = r(s,a) + \gamma \E_{s'\sim \mathbf{P}(s'|s,a)}\big[\max_{a'} Q^k(s',a')\big] \hspace{0.1in} \forall k \in \{1,2,\dots \}
\end{equation}
Further, each $Q^k$ can be represented using a neural network $N(s, a; \theta_k)$ with input $s$, output of dimension $A$, and parameters $\theta_k$, which are iteratively updated during training.

\subsection{DDQN}
DDQN approximates $Q$ values using two neural networks: A master network and a target network.  At every iteration $k$, the parameters of the master network are updated using Equation~\ref{eq:bellmanop}, where the inner maximization is replaced with the following function approximator:
\begin{equation}
\begin{aligned}
\max_{a'}Q^k(s,a') &=  N(s, \arg\max_{a'} N(s,a'; \theta_k); \theta'_k).
\end{aligned}
\end{equation}
The target network is then updated by setting $\theta'_k = \theta_k$ every $\tau$ iterations for some hyperparameter $\tau$. By using different networks to select and evaluate actions, DDQN attempts to address the issue of $Q$ value overestimation that may arise from taking the maximum of the estimated $Q$ values in Equation~\ref{eq:bellmanop} \citep{ddqn}.
\section{Methods}
\label{sec:methods}
In this work, we employ offline RL methods to improve the quality of treatment recommendations. We first describe a SOTA offline RL algorithm, CQL, that is used to generate a recommendation policy. Then, we explain our methodology to address the challenge of training CQL on retrospective datasets with action imbalance.  Finally, we describe an intuitive heuristic to enforce strict constraint satisfaction, and discuss how this is expected to impact the performance of CQL, or more generally of offline RL methods. 
\subsection{Conservative Q-Learning (CQL)}
A critical issue with directly applying Equation~\ref{eq:bellmanop} in the offline RL setting is that the $Q$ values of OOD actions may be overestimated.  To mitigate this, CQL combines the standard Q-learning objective of solving the Bellman equation with an additional regularization term that minimizes $Q$ values for OOD actions (i.e., large $Q$ values for OOD actions are penalized) \citep{cql}.  This gives rise to the objective
\begin{equation}
\label{eq:cqlobjective}
\begin{aligned}
\min_Q \max_{\mu} \quad & \alpha \cdot \big(\E_{s \sim \mathcal{D}, a \sim \mu(a|s)}\big[Q(s,a)\big]-E_{s \sim \mathcal{D}, a \sim \pi_{\beta}(a|s)}\big[Q(s,a)\big]\big) \\
\quad & + \frac{1}{2} \E_{(s,a,s')\sim \mathcal{D}}\big[(Q(s,a)-\mathcal{B}^k Q^k(s,a))^2\big] + \mathcal{R}(\mu).
\end{aligned}
\end{equation}
Here, $\mu$ is some distribution over the actions conditional on state, and $\mathcal{R}(\mu)$ is a regularization term.  Alternatives for $\mathcal{R}(\mu)$ proposed in \cite{cql} include: (i) Entropy, (ii) the negative of the KL divergence between $\mu$ and some given distribution, e.g., the learned policy at the previous iteration during model training, and (iii) a variant inspired by distributionally robust optimization, which penalizes the variance in the Q-function across actions and states.

Finally, $\alpha: \alpha > 0$ is a hyperparameter controlling the trade-off between the two objectives. While $\alpha$ can be tuned to learn less conservative policies, this alone may be insufficient to address the impact of severe action imbalance in the retrospective dataset $\mathcal{D}$. Hence, we propose a sampling method to reduce action imbalance. 

\subsection{CQL with sampling} \label{sec:cql-sampling}
CQL is designed to ensure that $Q$ value estimates for OOD actions are conservative. In settings like treatment optimization where the behavior policy exhibits large action imbalance, this may result in CQL predominantly recommending actions that are frequently recommended by the behavior policy, even when this is suboptimal. 
We thus propose to apply sampling approaches, which have been successfully used in many real-world classification problems to address class imbalance \citep{Undersamplingkubat, underoversamplingsolberg, overundersamplingling, adasyn, smote}, to similarly address the problem of action imbalance in offline RL.

Consider the following sampling procedure: Given action $a$, $1 \leq a \leq A$, suppose we sample with replacement a dataset $\mathcal{\hat{D}}$ from the set of historical transitions $\{(s_t, a, s_{t+1})\}$.  Denote the ratio of proportions of this set after sampling to before sampling as $w_a$ ($w_a > 1$: Action $a$ has been used less frequently in the retrospective dataset). Then $\mathcal{\hat{D}}$ has distribution 
\begin{equation}
\label{eq:sampleaconds}
\begin{aligned}
\Pr_{\mathcal{\hat{D}}}[a|s_t] &= w_a \Pr_{\mathcal{D}}[a|s_t].
\end{aligned}
\end{equation}
Since $\hat{D}$ is created by sampling with replacement from the set of historical transitions with action $a$, we also have, for each $a$:
\begin{equation}
\label{eq:samplessconda}
\begin{aligned}
\Pr_{\mathcal{\hat{D}}}[s_{t+1}|s_t,a] &= \Pr_{\mathcal{D}}[s_{t+1}|s_t,a],
\end{aligned}
\end{equation}
i.e., the transition probabilities are the same for $\mathcal{D}$ and $\mathcal{\hat{D}}$.  In the following, we will denote $\Pr_{\mathcal{\hat{D}}}[a|s_t]$ as $\pi_{\hat{\beta}}$.  To see how this approach affects the CQL recommendations, we write the CQL objective (Equation~\ref{eq:cqlobjective}) under the distribution $\mathcal{\hat{D}}$:
\begin{align}
& \E_{s_t \sim \mathcal{\hat{D}}} \Big[ \alpha \cdot \left(\E_{a_t \sim \mu\left(a_t|s_t\right)}\left[Q\left(s_t,a_t\right)\right]- \E_{a_t \sim \pi_{\hat{\beta}}\left(a_t|s_t\right)}\left[Q\left(s_t,a_t\right)\right]\right) \notag \\
&+  \frac{1}{2} \E_{\left(a_t,s_{t+1}\right)\sim \mathcal{\hat{D}}|s_t}\big[(Q\left(s_t,a_t\right)-\mathcal{B}^k Q^k\left(s_t,a_t\right))^2\big] \Big] + \mathcal{R}(\mu) \notag \\
= \quad & \E_{s_t \sim \mathcal{\hat{D}}} \Big[  \sum_a
 \alpha \cdot Q(s_t,at)( \mu(a|s_t)-w_a\pi_{\beta}(a|s_t))  \notag \\
&  + w_a \pi_{\beta}(a|s_t) \E_{s_{t+1} \sim \mathcal{D}|s_t,a} \big[ ( Q(s_t,a)-\mathcal{B}^{k}Q^k(s_t,a) )^2 \big] \Big] + \mathcal{R}(\mu). \label{eq:cqlsamplsummand2}
\end{align}
Here, the expectation is taken over the state $s_t$ under $\mathcal{\hat{D}}$.  Note that $s_t \sim \mathcal{D}$ is distinct from $s_t \sim \mathcal{\hat{D}}$ as the frequency of transitions with state $s_t$ is the sum of the frequencies of transitions with state $s_t$ and action $a$ across all $a$; This differs from $\mathcal{D}$ to $\mathcal{\hat{D}}$ due to Equation~\ref{eq:sampleaconds}.  The right hand side of Equation~\ref{eq:cqlsamplsummand2} then follows from our assumptions about the sampling process, where the first summand in the expectation follows from Equation~\ref{eq:sampleaconds} and the second summand is derived from Equation~\ref{eq:samplessconda}.  Thus when $w_a > 1$, the CQL overestimation term (first summand) decreases with sampling compared to without sampling, while the Bellman loss term (second summand) receives greater weight after sampling is applied, i.e., CQL becomes less conservative with respect to action $a$.  Conversely, when $w_a < 1$, the CQL overestimation term (first summand) increases with sampling compared to without sampling, while the Bellman loss term (second summand) receives smaller weight after sampling is applied, i.e., CQL becomes more conservative with respect to action $a$.

Motivated by this analysis, we propose combining CQL with sampling as follows: Instead of adopting $A$ hyperparameters $w_a$, which could be computationally costly to tune, we introduce a single hyperparameter $K$.  By tuning $K$ through grid-search, we can in turn tune all $w_a$ to reduce action imbalance.  Denoting the mean number of transitions per action in $\mathcal{D}$ as $\sigma$, we have: 1) Undersampling - for each action with more than $K \sigma$ transitions, sample $K \sigma$ transitions with replacement; 2) Oversampling - for each action with fewer than $K \sigma$ transitions, sample $K \sigma$ transitions with replacement; and 3) Under+oversampling - for each action, sample $\sigma$ transitions with replacement.  We then trained CQL on the sampled datasets. 


\subsection{Constraint satisfaction}
\label{sec:constraintsatisfaction}

The constraints from the diabetes setting take the form
\begin{equation} \label{eq:rloptconstrspecific}
\begin{aligned}
\max_{\pi} \quad & \E_{\pi}[\sum_{t=0}^T \gamma^t  R(s_t,a_t) | s_0, a_0 = 0] \\
\text{s.t.} \quad & \pi(s_t) \in A_F(s_t).
\end{aligned}
\end{equation}

For constrained optimization problems (see Equation~\ref{eq:rloptconstr}), direct application of CQL and DDQN does not guarantee constraint satisfaction during policy execution. For this special case, the learned policies of any value-based RL agent can be easily adapted to ensure strict constraint satisfaction. Given state $s_t$, denote the feasible set of actions as $A_F(s_t)$. For a given value-based RL agent $\pi_{RL}$, define the corresponding constrained policy as
\begin{equation}
\label{eq:constraintheuristic}
\begin{aligned}
\pi_{RL,c}(s_t) &= \arg \max_{a \in A_F(s_t)} Q^{\pi_{RL}}(s_t, a). 
\end{aligned}
\end{equation}
Here, $Q^{\pi_{RL}}(s_t,a)$ denotes the RL agent's estimated $Q$ value.  $\pi_{RL,c}(s_t)$ thus recommends the feasible action with the highest predicted $Q$ value. Then, the constrained and unconstrained recommended actions are the same when the latter is feasible.  

Intuitively, this means that if the RL agent's rate of constraint satisfaction is high, the optimality gap for the constrained recommendations should be close to the optimality gap for the unconstrained recommendations.  This is precisely expected to be the case for offline RL algorithms in treatment optimization settings. Since the SoC's rate of constraint satisfaction should be high (the SoC reflects domain experts' attention to safety considerations), and the offline RL agent's recommendations should not deviate too much from the SoC recommendations, the latter's constraint satisfaction rate should be high as well. Any performance guarantees for the offline RL agents' unconstrained recommendations should continue to the hold for constrained recommendations. This observation is supported by Property~\ref{constraintbound} below (proof in Appendix~\ref{sec:proofconstraints}).        

\begin{property}
\label{constraintbound}
Let $\pi^*$ be an unconstrained policy solving Equation~\ref{eq:rlopt}, and let $\pi_c^*$ be a constrained policy solving Equation~\ref{eq:rloptconstrspecific}.  Assuming that the reward $r(s,a)$ is bounded as $\underline{r} \leq r(s,a) \leq \overline{r}$, the optimality gap for $\pi_{RL,c}(s_t)$ can be bounded in terms of the optimality gap for $\pi_{RL}(s_t)$ as
\begin{equation}
\begin{aligned}
&\E[V^{\pi_c^*}(s_0)]-\E[V^{\pi_{RL,c}}(s_0)] \leq \frac{\overline{r}-\underline{r}}{1-\gamma} \sum_{t=0}^T \Pr[\exists t \text{ s.t. } \pi(s_t) \notin A_F(s_t)] + E[V^{\pi^*}(s_0)]-\E[V^{\pi_{RL}}(s_0)]. \\
\end{aligned}
\end{equation}
\end{property}
\section{Experimental Design}
\label{sec:experiments}

\subsection{Treatment optimization applications}

We conducted experiments comparing the recommendations of the different RL agents for two clinical applications: 1) type 2 diabetes, and 2) sepsis. We detail the tasks and datasets for these below.

\subsubsection{Diabetes treatment recommendation} \label{sec:diabetesdata} 
\textbf{Task.} We considered the problem of recommending antidiabetic treatment regimens to type 2 diabetes patients in an outpatient setting.  At each visit, the doctor prescribes a treatment regimen from among 13 options: 1) Maintain, 2) increase, or 3) decrease the dosages of previously prescribed drugs, or start a new drug from among the subgroups 4) acarbose, 5) DPP-4 inhibitors, 6) biguanides, 7) SGLT2 inhibitors, 8) sulphonylureas, 9) thiazolidinediones, 10) GLP-1 RAs, 11) long or intermediate acting insulins, 12) premixed insulins, and 13) rapid acting insulins.  The goal is to treat the patient's glycaeted haemoglobin (HbA\textsubscript{1c}) down to a target of 7\%, minimize the incidence of severe hypoglycemia, and reduce the incidence of complications such as heart failure \citep{ADAguidelines}.  This results in the following expression for reward at patient $i$'s visit at time $t$
\begin{equation}
\label{eq:diabetesreward}
\begin{aligned}
r_{i, t} &= 1\{s_{i, t+1}^{HbA_{1c}} \leq 7\} - 2s_{i, t+1}^{Hypo} - 4s_{i, t+1}^{Compl}
\end{aligned}
\end{equation}
where $s_{i, t}^{HbA_{1c}}$ is the HbA\textsubscript{1c}, $s_{i, t}^{Hypo}$ indicates the occurrence of hypoglycemia, and $s_{i, t}^{Compl}$ indicates the occurrence of complications or death at time $t$.  Similar reward functions are used in \cite{t2dmsdr} and \cite{t2dmkNN}.  For safety reasons, the treatment recommendation must also adhere strictly to clinical guidelines:
\begin{align}
C1: \quad & A^1_F(s_{i,t}) = \mathcal{A}_D \setminus \{6\} \text{ if } s_{i,t}^{eGFR} < 30 ml/min/1.73m^2 \label{eq:c1} \\
C2: \quad & A^2_F(s_{i,t}) = \mathcal{A}_D \setminus \{7\} \text{ if } s_{i,t}^{eGFR} < 45 ml/min/1.73m^2 \label{eq:c2} \\
C3: \quad & A^3_F(s_{i,t}) = \mathcal{A}_D \setminus \{10\} \text{ if } s_{i,t}^{pancr} = 1 \label{eq:c3} \\
C4: \quad & A^4_F(s_{i,t}) = \mathcal{A}_D \setminus \{8\} \text{ if } s_{i,t}^{age} > 70 \label{eq:c4}, 
\end{align}
where $s_{i,t}^{eGFR}$ is the estimated glomular filtration rate (eGFR),  $s_{i,t}^{pancr}$ is the incidence of pancreatitis, $s_{i,t}^{age}$ is the patient's age, and $\mathcal{A}_D = \{1, 2, \dots, 13\}$ is the set of actions.  The feasible set $A_F(s_{i,t})$ is then
\begin{equation}
A_F(s_{i,t}) = \bigcap_{j=1}^{4} A_F(s^{j}_{i,t}). 
\end{equation}

\textbf{Data.} We studied this problem using electronic medical records from outpatient prescription visits for type 2 diabetes patients within the Singapore Diabetes Registry \citep{sdr}. Our study was approved by the relevant Institutional Review Board with a waiver of informed consent. For each patient visit at time $t$, the state was defined by 55 variables describing the patient medical profile, including demographics (age, gender, ethnicity), physical measurements (heart rate, blood pressure, BMI), blood and urine laboratory data (HbA1c, fasting glucose, lipid panel, full blood counts, creatinine, estimated glomerular filtration rate, urine albumin/creatinine ratio), medical history (diabetes duration, utilization details, comorbidities and complications), and the previous visit prescription. We included prescription visits based on two inclusion criteria: (a) visit had at least one preceding prescription, and (b) visit had at least 1 HbA\textsubscript{1c} measurement within the past month and at least 1 eGFR reading within the past year.  This yielded 1,302,461 visits for 71,863 patients.

Among these visits, we observe significant action imbalance.  The most and least common action account for 64.0\% and 0.01\% of visits respectively. The most common action is "No change" and is prescribed in 51.2\% of visits where the patient's HbA\textsubscript{1c} is above the target of 7\%. This could be indicative of clinical inertia, where treatments are not intensified appropriately due to lack of time during the consultation, or lack of expertise in primary care settings \citep{shah2005clinicalinertia}. The action imbalance observed in this dataset, along with an over-representation of suboptimal actions, suggests that it is a good candidate for the methods proposed in Section~\ref{sec:cql-sampling}.

\subsubsection{Sepsis treatment recommendation} \label{sec:sepsisdata}
\textbf{Task.} We also considered the problem of treating ICU patients with sepsis.  At each 4 hour window $t$ during patient $i$'s ICU stay, the clinician administers fluids and/or vasopressors, each discretized into 5 volumetric categories.  The treatment is described by the tuple $(u,v), 1 \leq u,v \leq 5$, with 25 treatment options in total.  The goal being to prevent patient mortality, we model the treatment optimization problem with the following reward function:
\begin{equation} 
\label{eq:sepsisreward}
r_{i, t} = \begin{cases}
0 \text{ if } t < T \\
1 \text{ if } t = T \text{ and } s_{i,t}^{Mort} = 0 \\
-1 \text{ if } t = T \text{ and }  s_{i,t}^{Mort} = 1
\end{cases}
\end{equation}
where $s_{i,t}^{Mort}$ is the incidence of mortality within a 48 hour window of time $t$.  Similar reward functions have been used in \cite{fatemideadendstates} and \cite{aiclician}.

\textbf{Data.} We studied this problem using data on a cohort of sepsis patients generated from the publicly available MIMIC (Medical Information Mart for Intensive Care) - III dataset \citep{mimiciii} following \cite{fatemideadendstates}\footnote{We directly applied the associated code base provided in \url{https://github.com/microsoft/mimic_sepsis}.}. For each ICU stay and 4 hour window $t$, the state was defined by 44 variables describing the patient medical profile, including demographics (age, gender), physical measurements (heart rate, blood pressure, weight), and blood and urine laboratory data (glucose, creatinine). There were a total of 18923 unique ICU stays. We observed large action imbalance, with the most and least common actions accounting for 27.1\% and 1.9\% of visits respectively. The most common action was $(1,1)$, corresponding to the lowest possible dose ranges for IV fluids and vasopressors. Applying existing offline RL methods to learn from this dataset may thus result in recommendations for insufficiently intensive treatments.  

\subsection{Evaluations}
In the absence of data on counterfactuals, we considered evaluation techniques that use retrospective data.  

\subsubsection{Weighted Importance Sampling (WIS)}
We applied WIS, an off-policy evaluation technique widely used in the treatment optimization literature \citep{aiclician, sepsisraghu, sepsiswisAIM,raghusepsisb, sepsispeng}, to estimate the value of each RL agent's policy. We define the \textbf{WIS score} for an RL agent as follows. First, we define the importance ratios $\rho_{i, t}$ as the ratio of the likelihood of the RL agent policy $\pi_{RL}$ and the likelihood of the SoC policy $\pi_{Clin}$ selecting the SoC action $a_{i,t}$ given state $s_{i,t}$.  We then define trajectory-wise WIS estimators $V^{WIS}_i$ for trajectory $i, i = 1, \dots, N$, in terms of these importance ratios, and average $V^{WIS}_i$ across trajectories:
\begin{align}
\rho_{i, t} = \pi_{RL}(a_{i,t}|s_{i,t})/\pi_{Clin}(a_{i,t}|s_{i,t}) \hspace{0.5in} \nonumber \\
\rho_{i, 1:t} = \prod_{t'=1}^{t} \rho_{i, t'} ; \hspace{0.1in} w_{i,t} = \sum_{i=1}^N \rho_{i, 1:t}/N \hspace{0.5in} \\
V^{WIS}_i = \frac{\rho_{i, 1:t}}{w_{i,T}} \cdot \sum_{t=1}^T \gamma^{t-1}r_{i,t} ; \hspace{0.1in} WIS = \frac{1}{N}\sum_{i=1}^N V^{WIS}_i \nonumber 
\end{align}
This gives a biased but consistent estimator of the expected cumulative reward under the RL agent's policy \citep{wis}.

Following \cite{aiclician}, we estimated $\pi_{Clin}$ by training a multinomial logistic regression model with the one-hot-encoded selected action as the output, and the state as the features, then by taking the predicted probabilities for the different actions.  We estimated $\pi_{RL}$ by "softening" the RL agent policy, i.e., approximating it with a random policy $\pi^{WIS}_{RL}$ that selects from the non-optimal actions uniformly at random with some small probability.  Specifically,
\begin{equation} \label{eq:softening}
\pi^{WIS}_{RL}(s_{i,t}) = \begin{cases} 
a \text{ w.p. } \epsilon \text{ if } a^*_{i,t} \neq a \\
a^*_{i,t} \text{ w.p. } 1-\epsilon  \text{ otherwise.}
\end{cases}
\end{equation}
where $\epsilon$ ($0 < \epsilon \ll 1$) is a softening probability.  To ensure a fair comparison between the SoC and the RL agents, we applied softening to the SoC policy and calculated WIS estimates.  
\subsubsection{Additional metrics.}
We defined metrics of how well the RL agents' recommendations were aligned with clinical practice.

\textbf{Model Concordance Rate.} This is the fraction of visits where the RL policy's recommendation matches the SoC \citep{t2dmsdr, heparinddpg, heparin}:
\begin{equation}
\text{Model Concordance Rate} = \sum_{i,t}\mathbbm{1}\{\pi_{SoC}(s_{i,t}) = \pi_{RL}(s_{i,t})\}/\sum_{i} T_i. 
\end{equation}

\textbf{Appropriate Intensification Rate.} For the type 2 diabetes application, this is the fraction, out of visits with HbA1c over 7.0\%, where the RL agent recommends treatment intensification (i.e., increase dose or add new drug):
\begin{equation}
\text{Appr. Intens. Rate} = \sum_{i,t} \frac{\mathbbm{1}\{s_{i,t}^{HbA_{1c}} > 7.0\%,\pi_{RL}(s_{i,t}) \notin\{0,2\}\}}{\mathbbm{1}\{s_{i,t}^{HbA_{1c}} > 7.0\%\}}. 
\end{equation}

\textbf{Constraint Satisfaction Rate (CSR).} For the diabetes application, we defined the CSR for each constraint $j$, $j = 1$ to $4$ as the fraction of all visits where constraint $j$ applies such that the RL agent satisfies constraint $j$:
\begin{align*}
\text{CSR for constraint } j &= \sum_{i,t} \mathbbm{1}\{\pi_{RL}(s_{i,t} \in A^j_F(s_{i,t}))\}/\sum_{i} T_i. 
\end{align*}
As an illustrative example, for constraint 1, the CSR is the fraction, out of visits with eGFR under $30 ml/min/1.73m^2$, where the recommendation is not "add metformin."

\subsection{Implementation details}
\label{sec:implementationdetails}
We set three seeds, and for each seed, randomly split the patients in each dataset into training, validation, and test sets with the ratio 60:20:20.  We then used the \texttt{d3rlpy} library\footnote{\url{https://github.com/takuseno/d3rlpy}} in Python to train CQL and DDQN. For CQL, \texttt{d3rlpy} sets the regularization term $\mathcal{R}(\mu)$ in Equation~\ref{eq:cqlobjective} as entropy, i.e., $\mathcal{R}(\mu) = \E[-\log(\mu)]$.
This variant was also shown through ablation studies in \cite{cql} to generally outperform the other proposed variants.

For each seed, we trained (i) CQL with $\alpha \in$ [0.1, 0.5, 0.8, 0.9, 1.0]; (ii) CQL with undersampling and $K \in [0.4, 0.8, 1.2]$, CQL with oversampling and $K \in [0.4, 0.8]$, and CQL with under+oversampling; and (iii) DDQN.  Transition sampling was applied to the training set but not the validation and test sets.  These settings resulted in the sampling weights $w_a: 1 \leq 1 \leq A$ given in Table~\ref{tab:samplingstats} (Recall from Section~\ref{sec:cql-sampling} that $w_a$ denotes the frequency of transitions with action $a$ after sampling compared to before sampling). For both the diabetes and sepsis applications, under+oversampling gives rise to the greatest decrease in frequency (smallest $w_a$) and the greatest increase in frequency (largest $w_a$). Under+oversampling thus seems to be the most aggressive sampling approach for these settings, and undersampling the least aggressive approach.  Finally, we applied stratified random sampling to increase sample representativeness and to ensure that the empirical distributions of the sampled datasets are close to $\hat{\mathcal{D}}$ \citep{teddlie2007stratified}.  Details are in Appendix~\ref{sec:stratifiedsampling}. 
\begin{table}
  \caption{$\min_a \{w_a\}$ and $\max_a \{w_a\}$ with different sampling approaches for the diabetes and sepsis applications.  For undersampling and oversampling, ranges across various $K$ are provided.}
\label{tab:samplingstats}
\centering
  \begin{tabular}{ccccc}
    \toprule
     Application & Sampling approach & $\min_a \{w_a\}$ & $\max_a \{w_a\}$  \\
    \midrule
    Diabetes & Undersampling & 0.27-0.39 & 2.73-5.52 \\
& Oversampling & 0.66-0.82 & 254.89-406.87 \\
& Under+oversampling & 0.12 & 776.37 \\
Sepsis & Undersampling & 0.29-0.50 & 2.60-4.48 \\
& Oversampling & 0.67-0.85 & 8.05-12.67 \\
& Under+oversampling & 0.16 & 23.69 \\
\bottomrule
\end{tabular}
\end{table}

For each RL agent, we used multilayer perceptron architectures for the master and target Q networks and considered 2 different configurations: (i) 2 linear layers with 256 hidden units, (iii) 3 linear layers with 512 hidden units.  We set the batch size to 64, learning rate to $6.25\mathrm{e}{-5}$, and the target update interval to 8000 steps.  We used grid-search across all hyperparameter combinations to select the model with the highest WIS score on the validation sets, and evaluated this model. 
To compute WIS scores, we trained multinomial logistic regression models to approximate the clinician policy (details in Appendix~\ref{sec:wisdetails}), and applied a softening factor of 0.99. We also calculated WIS scores for the SoC using Equation~\ref{eq:softening}. 
\section{Results and Discussion}
\label{sec:results}
Results for the diabetes application are in Fig.~\ref{fig:treatmentrechistogram}, Table~\ref{tab:benchmarking}, Table~\ref{tab:benchmarkingconstr}, while results for the sepsis application are in Fig.~\ref{fig:mimicheatmaps_freq} and Table~\ref{tab:benchmarkingmimic}. We organize our findings into three subtopics: (i) Comparison of the offline RL method CQL with the off-policy RL method DDQN; (ii) Effect of sampling and regularization hyperparameter tuning on the performance of CQL; and (iii) Performance of the RL methods after the constraint satisfaction heuristic (Equation~\ref{eq:constraintheuristic}) is applied.

\subsection{Comparing CQL and DDQN}
Comparing the unconstrained recommendations of CQL with $\alpha = 1.0$ and DDQN for the diabetes treatment application, we first see from Fig.~\ref{fig:treatmentrechistogram} that the distribution across the treatment options is much closer to the SoC under CQL than under DDQN.  In particular, CQL is far more likely than DDQN to recommend no change, increasing dose, or decreasing dose, and far less likely than DDQN to recommend adding new medications. The model concordance rates in Table~\ref{tab:benchmarking} support this picture. Across 3 seeds, CQL exhibits a mean model concordance rate of 62.5\% with the SoC's recommendations, compared to only 1.5\% under DDQN.  As a result of the greater model concordance of the SoC with CQL than with DDQN, we see from Table~\ref{tab:benchmarkingconstr} that across the four constraints, the constraint satisfaction rates are  higher under CQL (between 98.7\% and 99.7\%) than under DDQN (between 88.1\% and 94.3\%); and are closer to the constraint satisfaction rates under the SoC (between 98.1\% and 100.0\%). Thus CQL generates recommendations that have far greater alignment to clinical practice than DDQN.

In terms of impact on health outcomes, Table~\ref{tab:benchmarking} shows that the WIS score averaged across 3 seeds is higher under CQL (3.653) than under DDQN (-3.055).  Both CQL and DDQN achieve greater WIS scores than the SoC (-6.741).  We can then conclude from Equation~\ref{eq:diabetesreward} that unlike the DDQN and SoC recommendations, the CQL recommendations are not expected to lead to complications and/or mortality on average. 

\begin{table*}
\small
  \caption{Comparison of treatment recommendations by the SoC, and the unconstrained and constrained recommendations by the different RL agents, for the diabetes application.  The mean and standard deviations of the results across 3 seeds are reported.  For each metric, the highest mean value across the RL agents is bolded, along with the associated standard deviation.}
  \label{tab:benchmarking}
  \centering
  \begin{tabular}{ccccccc}
    \toprule
    & \multicolumn{2}{c}{Model Concordance Rate} &\multicolumn{2}{c}{WIS score}& \multicolumn{2}{c}{Appr. Intensification Rate}\\
    Agent & Unconstrained & Constrained & Unconstrained & Constrained & Unconstrained & Constrained  \\
    \midrule
    SoC & 100.0\% $\pm$ 0.0\% & N.A. & -6.741 $\pm$ 3.549 & N.A. & 40.3\% $\pm$ 0.1\% & N.A. \\
    CQL with $\alpha=1.0$ & 62.5\% $\pm$ 0.3\% & 62.6\% $\pm$ 0.3\% & 3.653 $\pm$ 0.983 & 3.698 $\pm$ 0.947 & 37.8\% $\pm$ 0.7\% & 37.4\% $\pm$ 0.8\% \\
    CQL with $\alpha=0.9$ & 63.0\% $\pm$ 0.1\% & 63.1\% $\pm$ 0.1\% & 3.489 $\pm$ 1.647 & 3.522 $\pm$ 1.590 & 38.0\% $\pm$ 2.4\% & 37.7\% $\pm$ 2.4\% \\
    CQL with $\alpha=0.8$ & 63.3\% $\pm$ 0.2\% & 63.3\% $\pm$ 0.2\% & 3.795 $\pm$ 1.642 & 3.798 $\pm$ 1.635 & 39.1\% $\pm$ 1.0\% & 38.8\% $\pm$ 1.0\% \\
    CQL with $\alpha=0.5$ & \textbf{64.3\% $\pm$ 0.3}\% & \textbf{64.3\% $\pm$ 0.3\%} & 3.580 $\pm$ 0.820 & 3.663 $\pm$ 0.695 & 39.9\% $\pm$ 2.1\% & 39.6\% $\pm$ 2.0\%  \\
    CQL with $\alpha=0.1$ & 50.7\% $\pm$ 0.2\% & 50.9\% $\pm$ 0.2\% & 5.530 $\pm$ 1.018 & 5.496 $\pm$ 0.963 & 57.5\% $\pm$ 4.1\% & 57.1\% $\pm$ 4.0\% \\
    CQL with undersampling & 43.6\% $\pm$ 0.2\% & 44.1\% $\pm$ 0.2\% & 5.082 $\pm$ 1.062 & \textbf{5.763 $\pm$ 1.235} & 73.5\% $\pm$ 3.2\% & 66.8\% $\pm$ 5.9\%  \\
    CQL with oversampling & 61.5\% $\pm$ 0.5\% & 61.7\% $\pm$ 0.4\% & 4.070 $\pm$ 1.374 & 4.049 $\pm$ 1.342 & 40.8\% $\pm$ 1.6\% & 40.3\% $\pm$ 1.7\% \\
    CQL with under+oversampling & 41.3\% $\pm$ 0.5\% & 42.0\% $\pm$ 0.5\% & \textbf{5.721 $\pm$ 0.392} & 4.668 $\pm$ 1.692 & 76.1\% $\pm$ 10.1\% & 68.5\% $\pm$ 8.9\% \\
    DDQN & 1.5\% $\pm$ 0.0\% & 1.6\% $\pm$ 0.1\% & -3.055 $\pm$ 1.577 & -0.190 $\pm$ 3.152 & \textbf{97.5\% $\pm$ 0.3\%} & \textbf{97.1\% $\pm$ 0.3\%}  \\
  \bottomrule
\end{tabular}
\vspace{1ex}
\end{table*}

\begin{table*}
\caption{Mean and standard deviations across 3 seeds of the CSRs for the four constraints (Equations~\ref{eq:c1}-\ref{eq:c4}) in the diabetes application.  For each constraint, we bold the highest mean CSR(s) across RL agents, and the associated standard deviation(s).}
\label{tab:benchmarkingconstr}
  \begin{tabular}{ccccc}
    \toprule
    & \multicolumn{4}{c}{Constraint Satisfaction Rate (CSR)} \\
    Agent & C1 & C2 & C3 & C4  \\
    \midrule
    SoC & 99.6\% $\pm$ 0.1\% & 99.8\% $\pm$ 0.1\% & 100.0\% $\pm$ 0.0\% & 98.1\% $\pm$ 0.0\% \\
    CQL with $\alpha=1.0$ & 99.3\% $\pm$ 0.1\% & 99.6\% $\pm$ 0.0\% & 99.7\% $\pm$ 0.2\% & 98.7\% $\pm$ 0.2\% \\ 
    CQL with $\alpha=0.9$ & 99.4\% $\pm$ 0.2\% & 99.7\% $\pm$ 0.1\% & 99.9\% $\pm$ 0.1\% & 98.7\% $\pm$ 0.1\% \\
    CQL with $\alpha=0.8$ & 99.4\% $\pm$ 0.1\% & 99.7\% $\pm$ 0.1\% & \textbf{100.0\% $\pm$ 0.0\%} & 98.8\% $\pm$ 0.1\% \\
    CQL with $\alpha=0.5$ & \textbf{99.5\% $\pm$ 0.1\%} & \textbf{99.8\% $\pm$ 0.1\%} & 99.9\% $\pm$ 0.1\% & \textbf{98.9\% $\pm$ 0.2\%} \\
    CQL with $\alpha=0.1$ & 98.4\% $\pm$ 0.3\% & 99.3\% $\pm$ 0.1\% & 99.5\% $\pm$ 0.6\% & 98.4\% $\pm$ 0.2\% \\
    CQL with undersampling & 99.0\% $\pm$ 0.1\% & 99.4\% $\pm$ 0.1\% & \textbf{100.0\% $\pm$ 0.0\%} & 95.4\% $\pm$ 0.0\% \\
    CQL with oversampling & 99.2\% $\pm$ 0.4\% & 99.6\% $\pm$ 0.1\% & 99.9\% $\pm$ 0.2\% & 98.2\% $\pm$ 0.3\% \\
    CQL with under+oversampling & 97.3\% $\pm$ 0.5\% & 98.8\% $\pm$ 0.0\% & 99.9\% $\pm$ 0.1\% & 93.1\% $\pm$ 1.2\% \\
    DDQN & 90.0\% $\pm$ 2.9\% & 88.1\% $\pm$ 3.0\% & 94.3\% $\pm$ 2.6\% & 93.1\% $\pm$ 0.7\% \\
  \bottomrule
\end{tabular}
\end{table*}

\subsection{Effect of sampling and regularization hyperparameter tuning}

Next, we compare the recommendations of the various CQL agents for the diabetes and sepsis applications.  For sampling, results correspond to the agent with $K$ selected via grid-search.  Appendix~\ref{sec:appendixresults} provides a sensitivity analysis of how performance depends on $K$.

\begin{figure}[h]
  \centering \includegraphics[width=\linewidth]{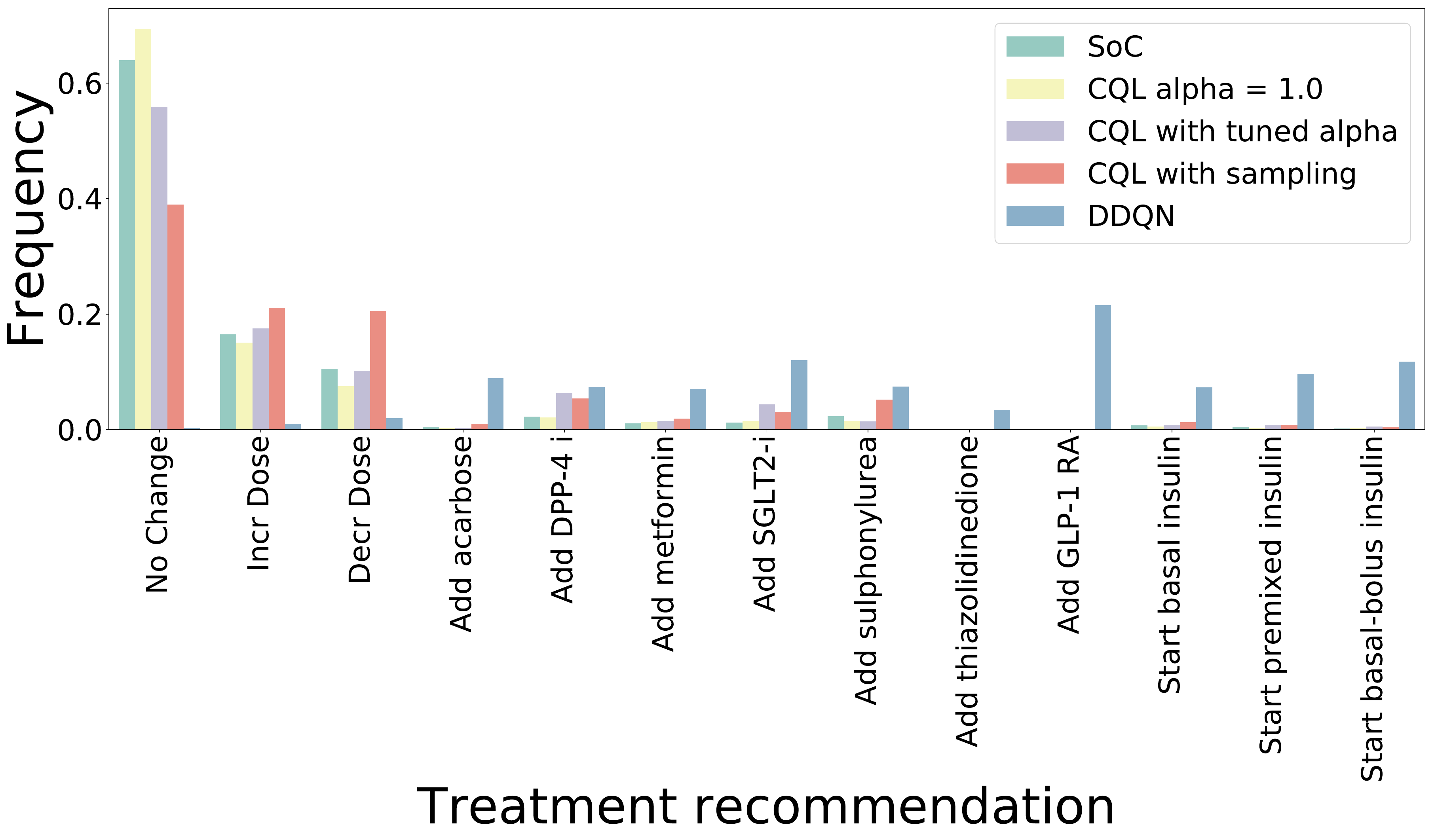}
  \caption{Distributions of unconstrained treatment recommendations under the various RL agents (as selected via grid-search) for the diabetes application.}
  \label{fig:treatmentrechistogram}
  \vspace{-0.1in}
\end{figure}

\textbf{Diabetes treatment application.}  Sampling and lowering $\alpha$ lead to greater divergence between the CQL and SoC recommendations:  Both contribute to a drop in the frequency of the no change recommendation and increases in the frequencies of each of the remaining treatment options (Fig.~\ref{fig:treatmentrechistogram}).  Similarly, the model concordance rate under CQL, as averaged over 3 seeds, decreases from 62.5\% with $\alpha = 1.0$ to 50.7\% with $\alpha = 0.1$, and to 43.6\% and 41.3\% with undersampling and under+oversampling respectively (Table~\ref{tab:benchmarking}).  Thus both regularization hyperparameter tuning and sampling have the expected effects of reducing the action imbalance of CQL's prescribed treatments.  At the same time, the CQL agents' recommendations are still more similar to the SoC than the DDQN baseline, i.e. as suggested in Section~\ref{sec:cql-sampling}, hyperparameter tuning and sampling continue to encourage conservativeness. This translates to generally higher rates of constraint satisfaction than under DDQN: Constraint satisfaction rates are between 98.4\% to 100.0\% for CQL with $\alpha < 1$, and between 93.1\% to 100.0\% for CQL with sampling (Table~\ref{tab:benchmarkingconstr}).

In terms of the optimality of the recommendations with hyperparameter tuning and sampling, Table~\ref{tab:benchmarking} shows an increase in the rate of appropriate treatment intensification from 37.8\% for CQL with $\alpha = 1.0$ to 57.5\% with $\alpha = 0.1$, to 73.5\% with undersampling, to 40.8\% with oversampling, and to 76.1\% with under+oversampling.  Compared to CQL with $\alpha = 1.0$, which has a WIS score of 3.653, the WIS scores are also higher for CQL with $\alpha = 0.1$ (5.530) and with $\alpha = 0.8$ (3.795); as well as for all the sampling schemes, with under+oversampling achieving the highest WIS score among all methods (5.721), followed by undersampling (5.082), then oversampling (4.070). Then, sampling and hyperparameter tuning both offer alternative means of improving the recommendations under CQL, with sampling outperforming hyperparameter tuning.

\begin{figure}[h]
  \centering \includegraphics[width=\linewidth]{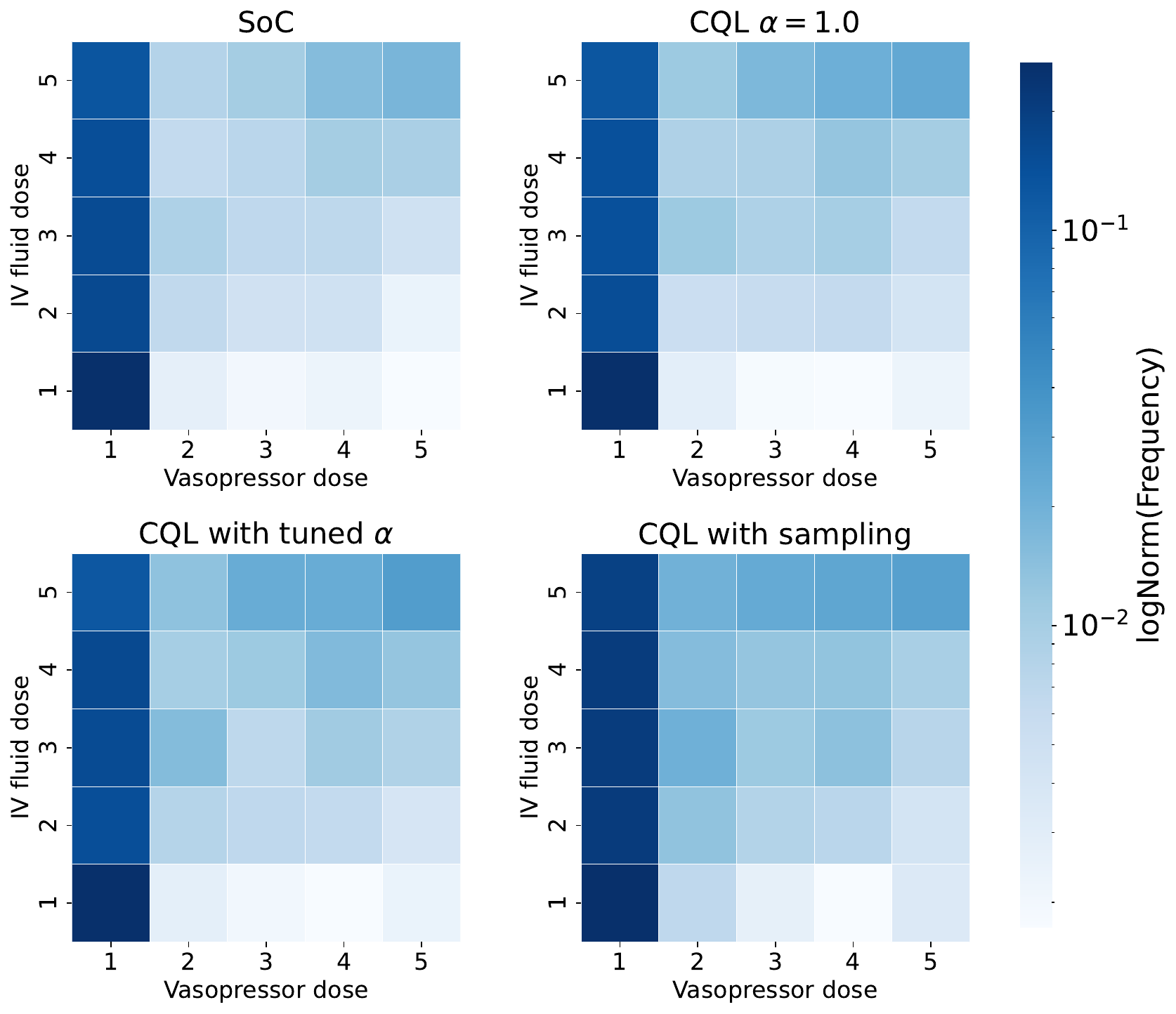}
  \caption{Distributions of treatment recommendations under SoC and the various RL agents (as selected via grid-search) on the sepsis dataset.}
  \label{fig:mimicheatmaps_freq}
  \vspace{-0.1in}
\end{figure}

\textbf{Sepsis treatment application.} For the sepsis application as well, sampling and lowering $\alpha$ lead to greater divergence between the CQL and SoC recommendations. Fig.~\ref{fig:mimicheatmaps_freq} shows that both reduce action imbalance, contributing to increases in the frequencies of the highest doses (top right section of each plot) of both IV fluids and vasopressors, as well as increases in the frequencies of higher IV fluid doses for the lowest vasopressor dose category.  The difference is more pronounced for sampling than for hyperparameter tuning, and indeed the model concordance rate in Table~\ref{tab:benchmarkingmimic} also decreases for all the sampling settings (from 30.1\% for $\alpha = 1.0$ to between 25.3\% and 28.5\% for the different sampling settings), while no decrease is observed for any $\alpha$, $\alpha < 1$.  

The WIS scores are also higher for CQL with $\alpha = 0.1$ (0.203), undersampling (0.191), oversampling (0.180), and under+oversampling (0.210), compared to CQL with $\alpha = 1.0$ (0.175).  CQL with under+oversampling attains the highest score.  From Equation~\ref{eq:sepsisreward}, this implies that survival rates are expected to increase for these CQL agents over CQL with $\alpha = 1.0$.  Thus, the finding that sampling and hyperparameter tuning can both improve the recommendations under CQL, with sampling outperforming hyperparameter tuning, generalizes from the diabetes to the sepsis application.
The relative performance of the different sampling approaches also generalizes to the sepsis application.  Under+oversampling outperforms undersampling, which in turn outperforms oversampling, in terms of WIS scores.  This is in contrast to the finding in the supervised learning setting that oversampling tends to outperform undersampling \citep{zha2022automated}.  A key difference between the supervised learning setting and our treatment optimization setting is that in the former, accuracy can suffer due to loss of information on majority class sample, while in the latter, it is more conservative and suboptimal actions that are overrepresented. Thus, undersampling is unlikely to be contributing to loss of information on the optimal action.

\begin{table}[h!]
  \caption{Comparison of treatment recommendations by the SoC, and the unconstrained recommendations of the different RL agents, on the sepsis dataset.  Metrics are Model Concordance Rate (MCR) and WIS score.  The mean and standard deviations of the results across 3 seeds are reported.  For each metric, the highest mean value across the RL agents is bolded, along with the associated standard deviation.}
  \centering
\label{tab:benchmarkingmimic}
  \begin{tabular}{cccc}
    \toprule
    Agent & MCR & WIS score \\
    \midrule
    SoC & 100.0\% $\pm$ 0.0\% & 0.095 $\pm$ 0.038  \\
    CQL with $\alpha=1.0$ & 30.1\% $\pm$ 0.2\% & 0.175 $\pm$ 0.018   \\
    CQL with $\alpha=0.9$ & 30.3\% $\pm$ 0.2\% & 0.135 $\pm$ 0.087  \\
    CQL with $\alpha=0.8$ & 30.4\% $\pm$ 0.2\% & 0.147 $\pm$ 0.056 \\
    CQL with $\alpha=0.5$ & \textbf{31.3\% $\pm$ 0.2\%} & 0.142 $\pm$ 0.051 \\
    CQL with $\alpha=0.1$ & \textbf{31.3\% $\pm$ 0.4\%} & 0.203 $\pm$ 0.075 \\
    CQL with undersampling & 28.0\% $\pm$ 0.5\% & 0.191 $\pm$ 0.053 \\
    CQL with oversampling & 28.5\% $\pm$ 0.1\% & 0.180 $\pm$ 0.044 \\
    CQL with under+oversampling & 25.3\% $\pm$ 0.2\% & \textbf{0.210 $\pm$ 0.035}  \\
  \bottomrule
  \vspace{-0.1in}
\end{tabular}
\vspace{-0.1in}
\end{table}

\vspace{-0.1in} 
\subsection{Evaluation of constrained recommendations}
With constrained recommendations, the distribution across actions did not change noticeably from Fig.~\ref{fig:treatmentrechistogram} (see Appendix~\ref{sec:appendixresults}). Model concordance rates for the unconstrained and constrained recommendations of each CQL agent are largely similar (Table~\ref{tab:benchmarking}).  This is due to the high constraint satisfaction rates for the unconstrained recommendations (Table~\ref{tab:benchmarkingconstr}). Consistent with the analysis in Section~\ref{sec:constraintsatisfaction}, the CQL agents' WIS scores experience only a small drop.

Our other findings also generalize to the constrained recommendations. Compared to DDQN, CQL with $\alpha = 1.0$ exhibits higher model concordance with the SoC (62.6\% vs. 1.6\%), and achieves a higher WIS score (3.698 vs. -0.190), implying that CQL's recommendations are more closely aligned with the SoC while also translating to improved expected health outcomes.  The WIS scores are higher for CQL with $\alpha = 0.8$ (3.798), $\alpha = 0.1$ (5.496), undersampling (5.763), oversampling (4.049), and under+oversampling (4.668), compared to CQL with $\alpha = 1.0$ (3.698). CQL with undersampling attains the highest score.  We conclude here as well that sampling and hyperparameter tuning can both improve over CQL recommendations, with sampling again outperforming hyperparameter tuning.
\section{Conclusion}
We have demonstrated that offline reinforcement learning (based on CQL) outperforms a popular deep off-policy RL method (DDQN) for a real-world diabetes treatment optimization application.  We found that offline RL recommendations are not only more closely aligned to clinical practice, but also translate to substantial improvements in expected health outcomes. Further, to address the common challenges of action imbalance encountered in real-world treatment optimization tasks, we devised a practical but theoretically grounded offline RL strategy for transition sampling of training data. Via extensive experiments for two real-world treatment optimization applications, we demonstrated improvements with this strategy over off-policy (DDQN) and offline (CQL) RL baselines in terms of expected health outcomes, as well as in terms of alignment of the recommendations with clinical practice guidelines.  Further, we showed theoretically and empirically that our results extend to when hard safety constraints are enforced via an intuitive heuristic. Our findings strongly suggest that offline RL should be chosen over off-policy RL for treatment optimization applications, as a means of enhancing safety and efficacy. Further, we highlight that transition sampling could find application in broader domains with critical safety considerations.

Future work could expand on our sampling approach by exploring more efficient automated sampling search strategies than grid-search.  Further, while we made efforts to conduct principled and extensive evaluations across a holistic set of metrics, inherent limitations with off-policy evaluations based on retrospective data should be acknowledged. Hence, future work could also include qualitative evaluations by clinician experts, who could rate the acceptability of the RL agents' recommendations.

\section*{Acknowledgements}
This research is supported by A*STAR, Singapore under its Industry Alignment Pre-Positioning Fund (Grant No. H19/01/a0/023 – Diabetes Clinic of the Future). We thank Sing Yi Chia and Nur Nasyitah Mohamed Salim for data preparation assistance, and the SingHealth Duke-NUS Institute of Precision Medicine for provisioning the computational infrastructure for this work.

\bibliographystyle{unsrtnat}
\bibliography{9_biblio}  

\appendix
\section{Theoretical analyses}

\subsection{Constraints: Proof of Property~\ref{constraintbound}.}
\label{sec:proofconstraints}
We can write the optimality gap of the RL model's policy after applying the heuristic, $\pi_{RL,c}$, as follows:
\begin{align}
&\E[V^{\pi_c^*}(s_0)]-\E[V^{\pi_{RL,c}}(s_0)] \notag \\
= \quad &\E[V^{\pi_c^*}(s_0)]-\E[V^{\pi^*}(s_0)] + \E[V^{\pi^*}(s_0)]- \E[V^{\pi_{RL}}(s_0)] + \E[V^{\pi_{RL}}(s_0)]- \E[V^{\pi_{RL,c}}(s_0)] \label{eq:constrgaplessthan0} \\
\leq \quad & \E[V^{\pi^*}(s_0)]-\E[V^{\pi_{RL}}(s_0)] \notag \\
&+ \E[V^{\pi_{RL}}(s_0)-V^{\pi_{RL,c}}(s_0) | \pi_{RL}(s_t) \in A_F(s_t) \forall t] \cdot \Pr[\pi_{RL}(s_t) \in A_F(s_t) \forall t] \notag \\
&+ \E[V^{\pi_{RL}}(s_0)-V^{\pi_{RL,c}}(s_0) | \exists t \text{ s.t. } \pi_{RL}(s_t) \notin A_F(s_t)] \cdot \Pr[\exists t \text{ s.t. } \pi_{RL}(s_t) \notin A_F(s_t)] \\
\leq \quad &  \E[V^{\pi^*}(s_0)]-\E[V^{\pi_{RL}}(s_0)] + \frac{\overline{r}-\underline{r}}{1-\gamma} \sum_{t=0}^T \Pr[\exists t \text{ s.t. } \pi_{RL}(s_t) \notin A_F(s_t)].
\end{align}
The first inequality comes from observing that the first summand in Equation~\ref{eq:constrgaplessthan0} is less than 0 by the optimality of $\pi^*$.  The second inequality comes from the assumption that $r$ is bounded, and applying the inequalities
\begin{equation*}
\frac{\underline{r}}{1-\gamma} \leq \E[V^{\pi_{RL}}(s_0)] \leq \frac{\overline{r}}{1-\gamma}.
\end{equation*}
This proves the theorem. $\square$

\section{Additional Implementation Details}

\subsection{Weighted Importance Sampling}
\label{sec:wisdetails}
To select a model for the clinician policy, we used the same seeds and train, test, and validation sets defined in Section~\ref{sec:implementationdetails}; and trained multinomial logistic regression models using the \texttt{sklearn} library in Python.  We set the one-hot-encoded SoC treatment recommendation as the dependent variable and the patient medical profile (state) as the independent variables.  We tested different regularization strengths, $C \in \{0.1, 1, 10, 10^2, 10^3\}$, and balancing of class weights to mitigate action imbalance (10 hyperparameter settings).  We selected the best model using the Brier score loss \cite{Brier1950VERIFICATIONOF} on the validation set.  Despite the action imbalance, models without balancing of class weights performed better. The results on the diabetes and sepsis test sets are reported in Table~\ref{tab:clinmodel}. 
\begin{table}[!htb]
  \caption{Performance of the best clinician model on the diabetes and sepsis test sets, in terms of means and standard deviations across 3 seeds for the Brier score and AUC.}
\centering
\label{tab:clinmodel}
  \begin{tabular}{ccc}
    \toprule
     Metric & Diabetes & Sepsis  \\
    \midrule
    Brier score loss & 0.487 $\pm$ 0.002 & 0.755 $\pm$ 0.003 \\
    AUC & 0.830 $\pm$ 0.003 & 0.851 $\pm$ 0.002  \\   
  \bottomrule
\end{tabular}
\end{table}

\subsection{Sampling}
\label{sec:stratifiedsampling}
We applied stratified random sampling based on gender, ethnicity, age, and diabetes duration for the diabetes treatment setting, and based on gender, age, and whether the ICU stay was a readmission for the sepsis treatment setting.  For continuous valued features, we binned values into 3 quantiles and stratified visits using the discretized values.

\section{Additional Results} \label{sec:appendixresults}
Here, we supplement the results in Section~\ref{sec:results}. 
For the diabetes and sepsis settings, Tables~\ref{tab:benchmarkingdiabetesbyK} and \ref{tab:benchmarkingmimicbyK} provide a sensitivity analysis of the performance of the performance of CQL with undersampling, oversampling, and under+oversampling for different $K$.

\begin{table*}
  \caption{Comparison of the unconstrained and constrained treatment recommendations by CQL with undersampling, CQL with oversampling, and CQL with under+oversampling, on the diabetes dataset, for different $K$.  Model Concordance Rate (MCR), WIS scores, and Appropriate Intensification Rates (AIR) are used as metrics. Results are the means and standard deviations across 3 seeds.}
  \label{tab:benchmarkingdiabetesbyK}\small
  \begin{tabular}{cccccccc}
    \toprule
    & & \multicolumn{2}{c}{MCR} &\multicolumn{2}{c}{WIS score}& \multicolumn{2}{c}{AIR}\\
    Sampling type & K & Unconstrained & Constrained & Unconstrained & Constrained & Unconstrained & Constrained  \\
    \midrule
    Undersampling & 0.4 & 36.6\% $\pm$ 0.8\% & 37.4\% $\pm$ 0.7\% & 5.082 $\pm$ 1.062 & 5.551 $\pm$ 1.191 & 73.5\% $\pm$ 3.2\% & 72.2\% $\pm$ 3.4\%  \\
    Undersampling & 0.8 & 41.4\% $\pm$ 6\% & 42.0\% $\pm$ 0.6\% & 5.738 $\pm$ 0.376 & 4.028 $\pm$ 2.176  & 63.6\% $\pm$ 0.5\% & 64.6\% $\pm$ 3.8\%  \\
    Undersampling & 1.2 & 43.6\% $\pm$ 0.2\% & 44.1\% $\pm$ 0.2\% & 5.297 $\pm$ 0.816 & 4.441 $\pm$ 1.675  & 60.9\% $\pm$ 0.3\% & 62.7\% $\pm$ 4.5\%  \\
    Oversampling & 0.4 & 61.9\% $\pm$ 2.6\% & 62.0\% $\pm$ 2.5\% & 5.037 $\pm$ 1.112 & 5.064 $\pm$ 1.177 & 36.8\% $\pm$ 1.7\% & 36.3\% $\pm$ 1.6\%  \\
    Oversampling & 0.8 & 61.5\% $\pm$ 0.5\% & 61.7\% $\pm$ 0.4\% & 4.070 $\pm$ 1.374 & 4.049 $\pm$ 1.342 & 40.8\% $\pm$ 1.6\% & 40.3\% $\pm$ 1.7\%  \\
    Under+oversampling & 1.0 & 41.3\% $\pm$ 0.5\% & 42.0\% $\pm$ 0.5\% & 5.721 $\pm$ 0.392 & 4.668 $\pm$ 1.692 & 76.1\% $\pm$ 10.1\% & 68.5\% $\pm$ 8.9\%  \\
  \bottomrule
\end{tabular}
\vspace{1ex}
\end{table*}

\begin{table*}[h!]
  \caption{Comparison of the unconstrained and constrained treatment recommendations by CQL with undersampling, CQL with oversampling, and CQL with under+oversampling, on the diabetes dataset, for different $K$. Model Concordance Rate (MCR) and WIS scores are used as metrics. Results are the means and standard deviations across 3 seeds.}
  \centering
\label{tab:benchmarkingmimicbyK}
  \begin{tabular}{ccccc}
    \toprule
    Sampling type & K & MCR & WIS score \\
    \midrule
    Undersampling & 0.4 & 25.9\% $\pm$ 0.2\% & 0.122 $\pm$ 0.102   \\
    Undersampling & 0.8 & 27.9\% $\pm$ 0.3\% & 0.142 $\pm$ 0.112  \\
    Undersampling & 1.2 & 28.0\% $\pm$ 0.5\% & 0.248 $\pm$ 0.082 \\
    Oversampling & 0.4 & 28.5\% $\pm$ 0.3\% & 0.210 $\pm$ 0.014 \\
    Oversampling & 0.8 & 28.5\% $\pm$ 0.1\% & 0.174 $\pm$ 0.037 \\
    Under+oversampling & 1.0 & 25.3\% $\pm$ 0.2\% & 0.210 $\pm$ 0.035 \\
  \bottomrule
\end{tabular}
\end{table*}

\begin{figure}[h]
  \centering \includegraphics[width=\linewidth]{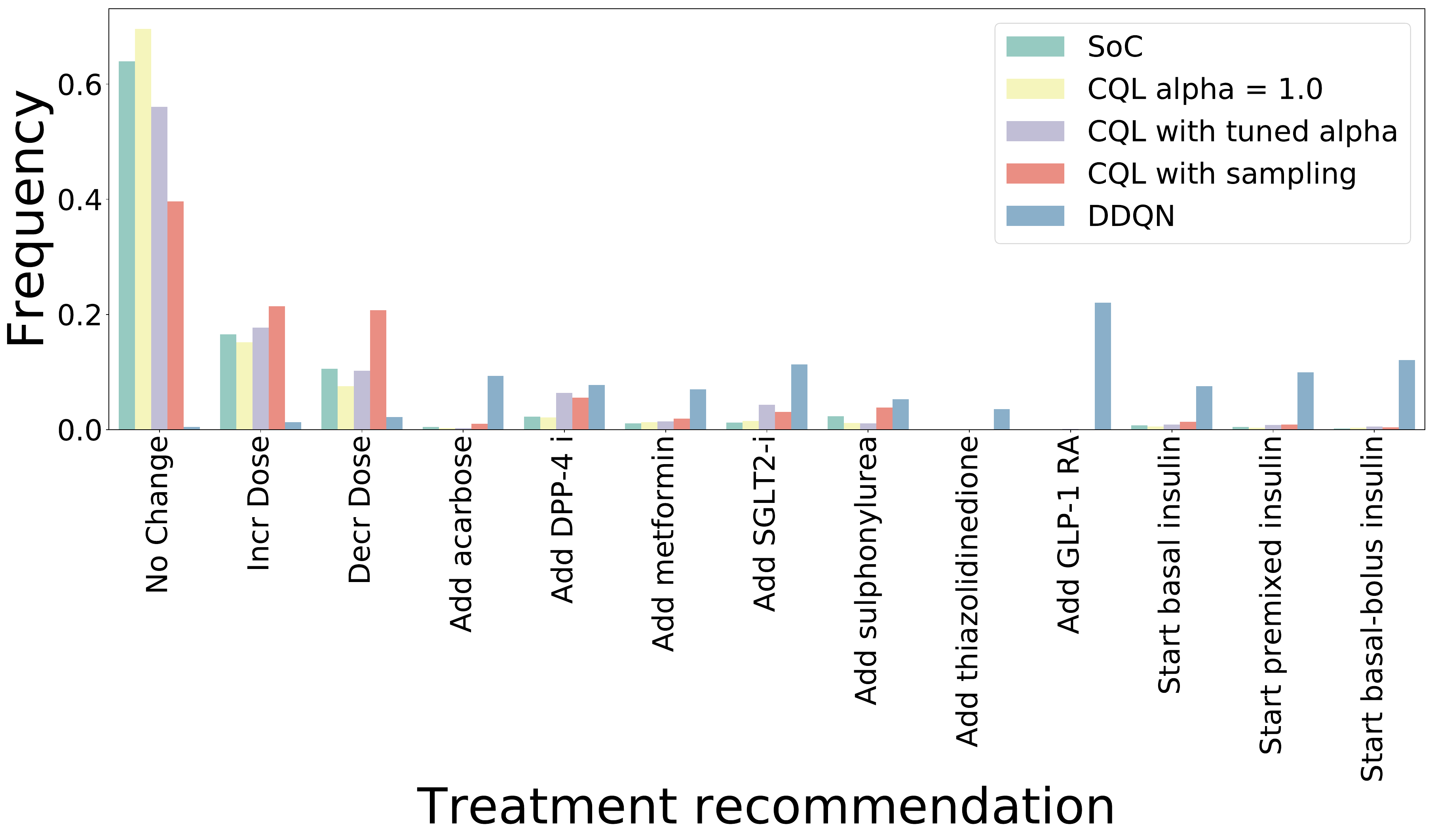}
  \caption{Distribution of the SoC treatment recommendations and the constrained treatment recommendations under CQL and DDQN for the diabetes setting.  The RL agents are selected via grid-search across all hyperparameter combinations, where the agent with the highest WIS score for constrained recommendations on the validation set is selected.}
  \label{fig:treatmentrechistogramconstrained}
\end{figure}

Finally, we complement Fig~\ref{fig:treatmentrechistogram}, which gives the distribution across treatment options of the RL models' unconstrained recommended treatments, with Fig.~\ref{fig:treatmentrechistogramconstrained}, which shows this distribution for the constrained recommendations. This distribution is not noticeably different from the distribution of unconstrained actions given in Fig~\ref{fig:treatmentrechistogram}, i.e., the constrained actions exhibit the same trends. All the CQL models' recommendations are more similar in distribution to the SoC than DDQN's recommendations.  Regularization parameter tuning and sampling lead to a greater divergence in distribution from the SoC recommendations compared to CQL with $\alpha = 1.0$.

\end{document}